\documentclass[letterpaper,10pt,conference]{ieeeconf}

\usepackage{amsmath}
\usepackage{graphicx}
\usepackage{subcaption}
\usepackage{multirow}
\usepackage{amsmath}
\usepackage{amssymb}

\usepackage{mathtools}
\usepackage{booktabs}
\usepackage{multirow}
\usepackage[export]{adjustbox}
\usepackage{array}

\IEEEoverridecommandlockouts                              

\makeatletter
\let\NAT@parse\undefined
\makeatother
\usepackage{verbatim}

\usepackage[colorlinks=true, allcolors=blue]{hyperref}

\title{\LARGE \bf PIVOT: Perception-aware Independent Viewpoint Online Optimization}

\author{Yuyang Chen$^{*,1}$, Shekoufeh Sadeghi$^{*,1}$, Charuvahan Adhivarahan$^{1}$, Elton Lemos$^{1}$,\\
Chen Wang$^{1}$, Sanjeev J. Koppal$^{2}$, and Karthik Dantu$^{1}$
\thanks{$^{*}$Equal contribution.}%
\thanks{$^{1}$Yuyang Chen, Shekoufeh Sadeghi, Charuvahan Adhivarahan, Elton Lemos, Chen Wang, and Karthik Dantu are with the Department of Computer Science and Engineering, University at Buffalo, Buffalo, NY 14260, USA.
{\tt\small {yuyangche, shekoufe, charuvah, eltonroq, cwx, kdantu}@buffalo.edu}}%
\thanks{$^{2}$Sanjeev J. Koppal is with the Department of Electrical and Computer Engineering, University of Florida, Gainesville, FL 32603, USA.
{\tt\small sjkoppal@ece.ufl.edu}}%
\thanks{This work has been submitted to the IEEE for possible publication. Copyright may be transferred without notice, after which this version may no longer be accessible.}%
}

\begin{document}
\maketitle

\begin{abstract}
A fundamental assumption in robotic perception is that the sensor's field of view (FoV) is fixed relative to the robot body. Motion-decoupled sensors, such as gimbal-mounted cameras and MEMS-based LiDARs, instead allow sensing direction to be controlled independently at runtime. This freedom creates a computational challenge: efficiently selecting useful viewing directions online in feature-dense environments. We propose PIVOT, a lightweight iterative method that optimizes sensor viewing direction along a fixed translation trajectory to maximize feature visibility. Under a conical FoV model, visibility depends only on the optical axis, yielding a two-degree-of-freedom optimization on the viewing sphere $\mathbb{S}^2$. Coordinate-free $\mathrm{SO}(3)$ exponential-map updates enable efficient continuous optimization without explicit angular parameterizations or exhaustive viewing-sphere search. Monte Carlo evaluations retain 98.1--99.6\% of brute-force visibility with a 76--85$\times$ speedup. Photorealistic simulation and real-world experiments further demonstrate improved visual localization robustness and practical viewpoint control on a quadruped robot.
\end{abstract}


\section{Introduction}
\begin{figure}[t]
    \centering
    \includegraphics[
        width=\linewidth
    ]{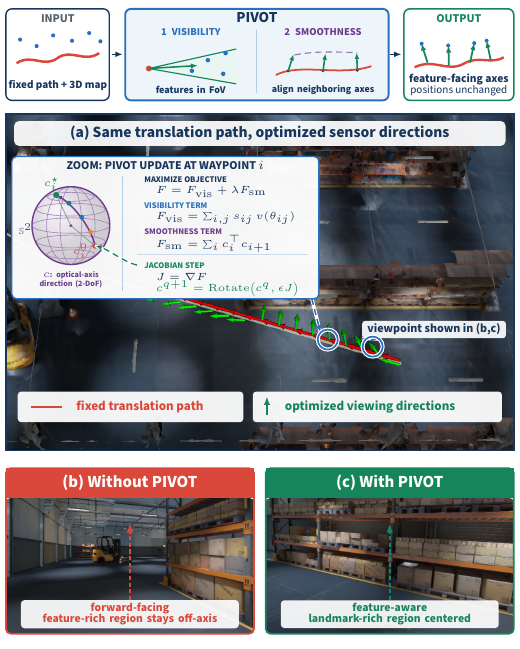}
    \caption{Given a fixed robot translation path and a 3D landmark map, PIVOT optimizes the sensor viewing directions on $\mathbb{S}^{2}$ to maximize feature visibility while encouraging smooth viewpoint changes. The inset in (a) summarizes the total objective and its Jacobian-driven on-manifold update. At the highlighted waypoint, a forward-facing camera leaves landmark-rich structure off-axis (b), whereas PIVOT rotates the optical axis to center it in the FoV (c).}
    \label{fig:main}
    \vspace{-15pt}
\end{figure}

Robotic autonomy depends not only on where a robot moves, but also on what its sensors can observe. In perception-driven tasks such as visual localization, inspection, tracking, exploration, and reconstruction, task-relevant features must remain within a limited field of view (FoV). This motivates \emph{active perception}~\cite{bajcsy1988active}, where sensing objectives are incorporated into planning to improve downstream perception~\cite{indelman2015planning,costante2016perception,alzugaray2017short,isler2016information,zhang2020fisher}.

Most perception-aware planning methods assume that sensing direction is coupled to the robot body, so changing the viewpoint requires changing the robot pose~\cite{zhang2018perception,zhang2020fisher}. This can be restrictive when robot motion is constrained by collision avoidance, dynamics, or task requirements~\cite{bonetto2021active,patel2019}. Motion-decoupled sensors, including gimbal-mounted cameras, pan--tilt units, and steerable MEMS-based LiDARs, instead allow sensing direction to be controlled independently of the robot body~\cite{liu2020real,jakobsen2005control,zou2006pan,wang2017ultra,wang2019large,wang2020low,tasneem2020adaptive,chen2024design}. Such sensing freedom has been exploited for UAV localization and active visual SLAM~\cite{patel2019,bonetto2021active}, and creates a complementary planning regime: when both robot motion and sensor pointing are free, they can be optimized jointly~\cite{papaioannou2023integrated}; when translation is prescribed or constrained, sensor pointing can be optimized independently along the given path.

This latter setting introduces a computational challenge. Visual localization and mapping may involve thousands to tens of thousands of landmarks, making exhaustive evaluation over candidate viewing directions expensive for high-rate operation. Existing approaches, including Fisher-information-based methods~\cite{zhang2019beyond,zhang2020fisher}, may couple sensing with robot motion, discretize candidate views, or rely on precomputed perception-quality representations~\cite{zhang2020fisher,wang2024active}. Our focus is lightweight online sensor-pointing optimization directly over dense or frequently updated feature maps, without exhaustive viewing-direction search or a separately precomputed perception-quality representation.

As shown in \autoref{fig:main}, given a pre-planned translation trajectory and a set of task-relevant 3D visual features, we compute a sequence of sensor viewing directions that maximizes feature visibility while discouraging abrupt changes between consecutive viewpoints. Under our conical FoV model, visibility depends only on the optical-axis direction, yielding a two-DoF optimization on \(S^2\). We derive a gradient-based update for a smooth FoV visibility objective and implement the update through \(SO(3)\) exponential-map rotations, enabling continuous viewpoint optimization without explicit angular parameterization or exhaustive search over viewing directions. The formulation operates directly on the feature point cloud, naturally supports feature-weighted visibility objectives, and extends to trajectory-level optimization through a smoothness term that promotes continuous sensor pointing and temporal view overlap.

Because the optimizer can be invoked repeatedly as the robot moves, it can support viewpoint adaptation as landmarks are added or task-relevant features change. Although we evaluate the method using known feature maps, its millisecond-scale runtime makes it suitable for integration with receding-horizon navigation, incremental mapping, active localization recovery, and inspection systems using independently actuated cameras or steerable depth sensors~\cite{chen2024design,tasneem2020adaptive}.

We evaluate the method through synthetic Monte Carlo experiments, photorealistic visual-localization simulations, and real-world experiments on a quadruped robot equipped with a pan--tilt camera. The proposed optimizer preserves near-brute-force visibility quality while reducing per-pose computation by up to two orders of magnitude, and active sensor pointing improves visual localization by keeping informative features within the camera FoV.

The main contributions of this paper are:
\begin{itemize}
    \item We formulate active FoV control for motion-decoupled sensors as an online viewing-direction optimization problem on \(S^2\), separating sensor pointing from robot translation planning.
    
    \item We derive a smooth feature-visibility objective and an on-manifold gradient update, implemented through \(SO(3)\) exponential-map rotations, without explicit angular parameterization or exhaustive candidate-view search.
    
    \item We extend the single-pose optimizer to trajectory-level viewpoint optimization using a smoothness objective for continuous sensor pointing and temporal view overlap.
    
    \item We demonstrate near-brute-force visibility at substantially lower computational cost, with validation in photorealistic simulation and on a quadruped robot.
\end{itemize}

\section{Related Work}

Active perception has long been used to improve localization and mapping by incorporating perception quality into planning. Prior work in perception-aware navigation and active SLAM plans robot motion using objectives related to localization uncertainty, scene observability, or SLAM quality~\cite{costante2016perception,alzugaray2017short,zhang2018perception,zhao2022perception}. Costante et al.~\cite{costante2016perception} incorporate photometric and geometric scene information into path planning to reduce localization uncertainty, while Alzugaray et al.~\cite{alzugaray2017short} perform short-term UAV planning with monocular-inertial SLAM in the loop. These approaches largely consider body-coupled sensing, where changes in viewing direction are realized through robot motion~\cite{zhang2018perception,zhou2021raptor}. Independently actuated sensors instead expose sensor pointing as a separate control variable, enabling viewpoint adaptation without requiring corresponding changes in robot motion. This decoupling has been used for active visual SLAM~\cite{bonetto2021active}, robust route-following localization with gimballed cameras~\cite{patel2019}, and hierarchical sensor planning~\cite{parandekar2024informative}.

A closely related line of work uses information-theoretic objectives for active visual localization. Zhang and Scaramuzza introduced the Fisher Information Field (FIF), which enables efficient online perception-aware planning through a precomputed representation of viewpoint-dependent information~\cite{zhang2019beyond,zhang2020fisher}. While FIF provides microsecond-scale online queries, constructing the field requires preprocessing, with reported build times ranging from several seconds to tens of seconds~\cite{zhang2019beyond,zhang2020fisher}. Although FIF supports incremental updates as landmarks are added or removed, it still maintains an intermediate perception-quality representation whose construction and update introduce additional map-management overhead. Wang et al.~\cite{wang2024active} more recently proposed active view planning using a continuous information model, but likewise rely on an intermediate global representation. In contrast, our approach operates directly on the current feature point cloud and requires no separate perception-quality field construction.

Recent work also demonstrates the practical value of independently
steerable sensors. Bonetto et al.~\cite{bonetto2021active} optimize an
independently rotating camera for active visual SLAM, while Patel
et al.~\cite{patel2019} and Spurny et al.~\cite{spurny2022active}
show that active gimbal control can improve UAV localization
robustness. More recently, Li et al.~\cite{li2025actloc} proposed
ActLoc, which learns viewpoint-dependent localization quality over
camera pitch and yaw at arbitrary 3D waypoints and integrates this
prediction into path planning. Our focus is complementary:
lightweight online optimization of sensor viewing direction directly
over dense feature maps along a prescribed translation trajectory,
without requiring a learned viewpoint-quality model.

Related viewpoint-optimization methods also address settings with different computational requirements. N\"ageli et al.~\cite{nageli2017real} perform real-time viewpoint optimization for aerial videography with a small number of cinematic targets, whereas visual localization may involve thousands to tens of thousands of features. Gemerek et al.~\cite{gemerek2022directional} jointly plan mobile-sensor motion while accounting for visibility, obstacles, robot geometry, and kinodynamic constraints. In our setting, translation is already given and only sensor pointing is optimized online, removing waypoint and path selection from the optimization problem. Watterson et al.~\cite{watterson2018trajectory} provide general tools for trajectory optimization on manifolds, including $SO(3)$; our use of rotational updates is complementary, with the contribution here centered on dense feature-visibility optimization for motion-decoupled sensing. By restricting online optimization to sensor pointing along a pre-planned path, PIVOT avoids exhaustive viewing-direction search and separate perception-field construction while scaling to dense feature maps in real time (\mbox{\autoref{sec:evalsim}}).

\section{Problem Formulation}
\label{sec:problem}

We consider a robot equipped with a motion-decoupled visual sensor,
such as a pan--tilt or gimbal-mounted camera. The robot follows a fixed
pre-planned translation trajectory while sensor pointing is optimized
to keep task-relevant visual features inside its limited field of view
(FoV).

Let the sensor-center trajectory in the world frame be
\begin{equation}
    \mathcal{T}
    =
    \{\boldsymbol{t}_0^w,\ldots,\boldsymbol{t}_N^w\},
    \qquad
    \boldsymbol{t}_i^w \in \mathbb{R}^3,
    \label{eq:translation_path}
\end{equation}
and the visual-feature set be
\begin{equation}
    \mathcal{P}
    =
    \{\boldsymbol{p}_1^w,\ldots,\boldsymbol{p}_M^w\},
    \qquad
    \boldsymbol{p}_j^w \in \mathbb{R}^3.
    \label{eq:feature_set}
\end{equation}

At waypoint $i$, the camera-to-world rotation is
$\boldsymbol{R}_i^w\in SO(3)$. With camera-frame optical axis
$\boldsymbol{e}_3=[0,0,1]^T$, the world-frame viewing direction is
\begin{equation}
    \boldsymbol{c}_i
    =
    \boldsymbol{R}_i^w\boldsymbol{e}_3
    \in S^2.
    \label{eq:view_direction}
\end{equation}
Under the conical FoV model, visibility depends only on
$\boldsymbol{c}_i$ and is invariant to rotation about the optical axis.
Thus, the effective optimization has two rotational degrees of freedom
on $S^2$, while $\boldsymbol{R}_i^w$ is retained for coordinate-free
rotational updates. The FoV half-angle is $\alpha$.

For feature $\boldsymbol{p}_j^w$ relative to waypoint
$\boldsymbol{t}_i^w$, the normalized world-frame bearing is
\begin{equation}
    \boldsymbol{k}_{ij}
    =
    \frac{\boldsymbol{p}_j^w-\boldsymbol{t}_i^w}
         {\|\boldsymbol{p}_j^w-\boldsymbol{t}_i^w\|}.
    \label{eq:bearing_world}
\end{equation}
The angle $\theta_{ij}$ between the feature bearing and optical axis
satisfies
\begin{equation}
    \cos\theta_{ij}
    =
    \boldsymbol{k}_{ij}^T\boldsymbol{c}_i
    =
    \boldsymbol{k}_{ij}^T
    \boldsymbol{R}_i^w\boldsymbol{e}_3.
    \label{eq:cos_theta_world}
\end{equation}

\begin{figure}[t]
\centering
\begin{subfigure}[t]{.44\linewidth}
    \centering
    \includegraphics[width=\linewidth, trim={0pt 0pt 0pt 8pt}, clip]{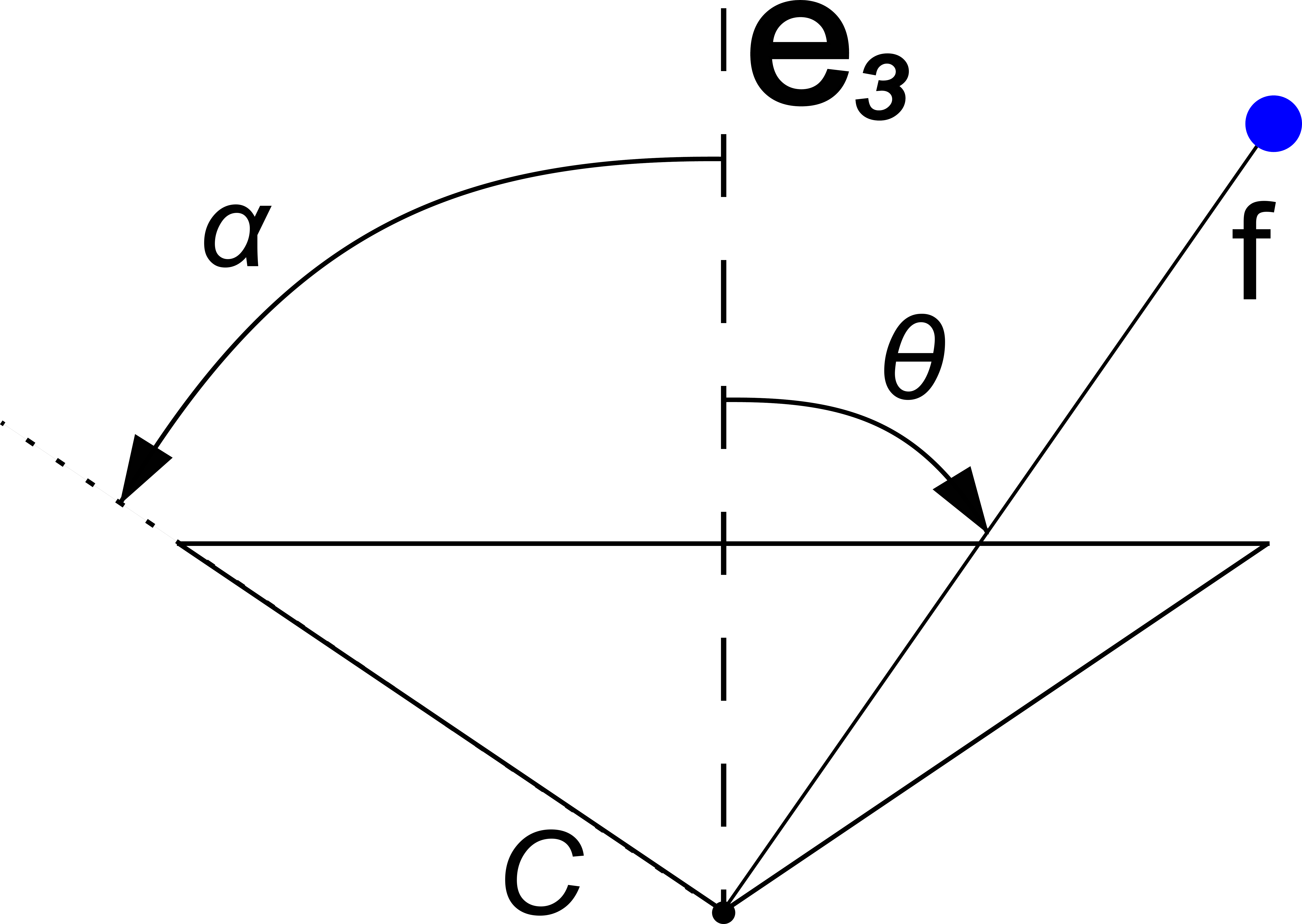}
    \label{fig:FOV_f_drawing}
\end{subfigure}
\hfill
\begin{subfigure}[t]{.54\linewidth}
    \centering
    \includegraphics[width=\linewidth, trim={2pt 5pt 5pt 0pt}, clip]{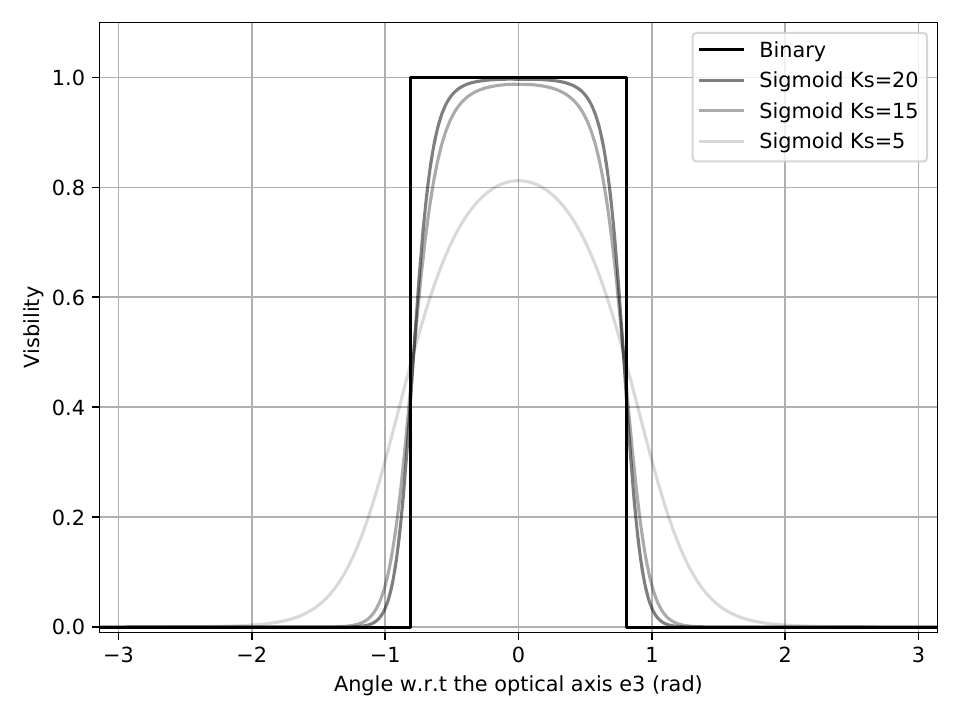}
    \label{fig:binary_sigmoid}
\end{subfigure}
\caption{Visibility model adapted from\cite{zhang2020fisher}.
(a) Conical FoV geometry, where $\alpha$ is the FoV half-angle,
$\boldsymbol{f}$ is a visual feature, $C$ is the sensor center, and
$\boldsymbol{e}_3$ is the optical axis.
(b) Binary and sigmoid visibility functions for varying $k_s$, with
$\alpha=45^\circ$.}
\label{fig:visibility}
\vspace{-15pt}
\end{figure}
As illustrated in~\autoref{fig:visibility}, a feature lies inside the FoV when $\theta_{ij}\leq\alpha$. We use the
smooth visibility approximation
\begin{equation}
    v(\theta_{ij})
    =
    \frac{1}
         {1+e^{-k_s(\cos\theta_{ij}-\cos\alpha)}},
    \label{eq:smooth_visibility}
\end{equation}
where $k_s>0$ controls the transition sharpness near the FoV boundary.
The visibility objective at waypoint $i$ is
\begin{equation}
    F_i(\boldsymbol{R}_i^w)
    =
    \sum_{j\in\mathcal{V}_i}
    s_{ij}\,v(\theta_{ij}),
    \label{eq:single_pose_visibility_objective}
\end{equation}
where $\mathcal{V}_i\subseteq\{1,\ldots,M\}$ is the retained landmark
set and $s_{ij}\geq0$ is the feature weight.

Independent waypoint optimization can produce abrupt viewing-direction
changes. We therefore optimize
\begin{equation}
    \max_{\boldsymbol{R}_0^w,\ldots,\boldsymbol{R}_N^w\in SO(3)}
    \sum_{i=0}^{N}
    F_i(\boldsymbol{R}_i^w)
    +
    \lambda
    \sum_{i=0}^{N-1}
    \bigl(\boldsymbol{R}_{i+1}^w\boldsymbol{e}_3\bigr)^T
    \bigl(\boldsymbol{R}_{i}^w\boldsymbol{e}_3\bigr),
    \label{eq:trajectory_objective}
\end{equation}
where $\lambda\geq0$ balances visibility and smooth sensor pointing.
The second term encourages alignment between consecutive optical axes,
promoting temporal view overlap. Since both terms depend on
$\boldsymbol{R}_i^w$ only through
$\boldsymbol{R}_i^w\boldsymbol{e}_3$, the objective is equivalently
defined over viewing directions on $S^2$. Translation remains fixed.

\section{Approach}
\label{sec:approach}

We solve~\autoref{eq:trajectory_objective} using iterative on-manifold
updates. Rotations are represented by $\boldsymbol{R}\in SO(3)$, while
the objective depends only on
$\boldsymbol{R}\boldsymbol{e}_3\in S^2$. At each iteration, we
parameterize a local perturbation in $so(3)$ and apply the corresponding
incremental rotation through the matrix exponential, avoiding explicit
angular parameterization and viewing-sphere discretization.

\subsection{On-Manifold Viewing-Direction Optimization}
\label{subsec:FOV_Optimization}

Dropping the waypoint index, let $\bar{\boldsymbol{R}}^w$ denote the
current camera rotation and define
\begin{equation}
    \boldsymbol{k}_j \equiv \boldsymbol{k}_{ij},
    \qquad
    s_j \equiv s_{ij},
    \qquad
    \mathcal{V}\equiv\mathcal{V}_i,
    \qquad
    \boldsymbol{c}=\bar{\boldsymbol{R}}^w\boldsymbol{e}_3.
    \label{eq:k_c_definitions}
\end{equation}

We parameterize a local perturbation by
$\boldsymbol{\omega}\in\mathbb{R}^3$, with
$\boldsymbol{\omega}_{\wedge}\in so(3)$ its corresponding
skew-symmetric Lie-algebra element. The exponential map converts this
perturbation to an incremental rotation. Using a left perturbation,
\begin{equation}
    \boldsymbol{R}^w(\boldsymbol{\omega})
    =
    e^{\boldsymbol{\omega}_{\wedge}}
    \bar{\boldsymbol{R}}^w,
    \label{eq:lie_update}
\end{equation}
where $\boldsymbol{\omega}$ is expressed in the world frame and
$\boldsymbol{R}^w(\boldsymbol{0})=\bar{\boldsymbol{R}}^w$.

The perturbed feature-angle cosine is
\begin{equation}
    \cos\theta_j(\boldsymbol{\omega})
    =
    \boldsymbol{k}_j^T
    e^{\boldsymbol{\omega}_{\wedge}}
    \boldsymbol{c}.
    \label{eq:cos_lie}
\end{equation}
Using
$e^{\boldsymbol{\omega}_{\wedge}}
\approx\boldsymbol{I}+\boldsymbol{\omega}_{\wedge}$,
\begin{align}
    \cos\theta_j(\boldsymbol{\omega})
    &\approx
    \boldsymbol{k}_j^T\boldsymbol{c}
    +
    \boldsymbol{\omega}^T
    (\boldsymbol{c}\times\boldsymbol{k}_j),
    \label{eq:cos_taylor}
\end{align}
giving
\begin{equation}
    \boldsymbol{J}_j
    =
    \left.
    \frac{\partial\cos\theta_j}
         {\partial\boldsymbol{\omega}}
    \right|_{\boldsymbol{\omega}=\boldsymbol{0}}
    =
    \boldsymbol{c}\times\boldsymbol{k}_j.
    \label{eq:jacobian}
\end{equation}
Since $\boldsymbol{J}_j^T\boldsymbol{c}=0$, the gradient lies in
$T_{\boldsymbol{c}}S^2$ and changes only the optical-axis direction.

Using~\autoref{eq:smooth_visibility}, the perturbed objective is
\begin{equation}
    F\!\bigl(\boldsymbol{R}^w(\boldsymbol{\omega})\bigr)
    =
    \sum_{j\in\mathcal{V}}
    s_j\,v(\theta_j(\boldsymbol{\omega})).
    \label{eq:objective}
\end{equation}
At the current iterate, define
\begin{equation}
    u_j=\boldsymbol{k}_j^T\boldsymbol{c},
    \qquad
    a_j=-k_s(u_j-\cos\alpha).
    \label{eq:u_a_definitions}
\end{equation}
The visibility-gradient contribution from feature $j$ is
\begin{equation}
    \boldsymbol{J}_{\text{vis},j}
    =
    s_j k_s
    \frac{e^{a_j}}{(1+e^{a_j})^2}
    \boldsymbol{J}_j,
    \label{eq:j_vis_single}
\end{equation}
and the full visibility gradient is
\begin{equation}
    \boldsymbol{J}_{\text{vis}}
    =
    \sum_{j\in\mathcal{V}}
    \boldsymbol{J}_{\text{vis},j}.
    \label{eq:j_vis}
\end{equation}
The single-waypoint update is
\begin{equation}
    \boldsymbol{R}_{q+1}^w
    =
    e^{\left(\epsilon\,\boldsymbol{J}_{\text{vis}}\right)_{\wedge}}
    \boldsymbol{R}_{q}^w,
    \qquad
    \epsilon=\frac{1}{n\pi},
    \label{eq:optimization_loop}
\end{equation}
where $q$ is the iteration index and $n=|\mathcal{V}|$. In
implementation, the tangent-space step is clipped to prevent
overshooting.

\subsection{Trajectory Optimization}
\label{subsec:Trajectory_Optimization}

For trajectory-level optimization, retained landmarks are weighted by
distance:
\begin{equation}
    s_{ij}
    =
    1-
    \frac{d_{ij}-d_i^{\min}}
    {\max\!\left(10^{-6},d_i^{\max}-d_i^{\min}\right)},
    \qquad
    d_{ij}
    =
    \|\boldsymbol{p}_j^w-\boldsymbol{t}_i^w\|,
    \label{eq:distance_weight}
\end{equation}
where
$d_i^{\min}=\min_{j\in\mathcal{V}_i}d_{ij}$ and
$d_i^{\max}=\max_{j\in\mathcal{V}_i}d_{ij}$.

Let
\begin{equation}
    \boldsymbol{c}_{i,q}
    =
    \boldsymbol{R}_{i,q}^w\boldsymbol{e}_3.
    \label{eq:trajectory_view_direction}
\end{equation}
The smoothness term is
\begin{equation}
    \sum_{i=0}^{N-1}
    \boldsymbol{c}_{i+1,q}^{T}\boldsymbol{c}_{i,q}.
    \label{eq:objective_trajectory}
\end{equation}
For an interior waypoint, its local gradient is
\begin{equation}
    \boldsymbol{J}_{\text{smooth},i}
    =
    \boldsymbol{c}_{i,q}
    \times
    \left(
    \boldsymbol{c}_{i-1,q}
    +
    \boldsymbol{c}_{i+1,q}
    \right),
    \quad 1\leq i\leq N-1,
    \label{eq:j_smooth}
\end{equation}
with only the available neighbor used at the endpoints. The combined
update direction is
\begin{equation}
    \boldsymbol{J}_i
    =
    \boldsymbol{J}_{\text{vis},i}
    +
    \lambda\,\boldsymbol{J}_{\text{smooth},i},
    \label{eq:combined_jacobian}
\end{equation}
and the trajectory update is
\begin{equation}
    \boldsymbol{R}_{i,q+1}^w
    =
    e^{\left(\epsilon\,\boldsymbol{J}_i\right)_{\wedge}}
    \boldsymbol{R}_{i,q}^w.
    \label{eq:trajectory_update}
\end{equation}
Only sensor pointing is optimized; translation remains fixed.

\subsection{Practical Optimization Enhancements}
\label{subsec:practical_enhancements}

To improve convergence and robustness, we use three practical
modifications.

\par\noindent\textbf{Multi-start initialization:}
Candidate viewing directions are generated toward the feature-map
centroid and from orthogonal and tangent directions on the viewing
sphere. They are ranked by the number of features inside the FoV,
near-duplicates are pruned, and only the top candidates undergo
on-manifold optimization, with early stopping when improvement
saturates.

\par\noindent\textbf{Scheduled FoV and dynamic $k_s$:}
When most features lie outside the initial FoV, sigmoid gradients can
be weak. Early iterations therefore use a wider effective half-angle,
up to $90^\circ$, and progressively narrow it to the target $\alpha$.
During this schedule, $k_s$ is recomputed from the current FoV and a
target angular transition width to maintain a consistent soft boundary.

\par\noindent\textbf{Joint update and step control:}
Trajectory updates use the combined gradient
in~\autoref{eq:combined_jacobian}. The update is normalized by local
feature count, clipped to prevent overshooting, and scaled adaptively
according to how often the clipping bound is reached.

\section{Evaluation in Simulation}
\label{sec:evalsim}

\begin{figure}[t]
    \centering
    \begin{subfigure}[b]{0.48\columnwidth}
        \centering
        \includegraphics[width=\linewidth,trim={60pt 40pt 30pt 40pt},clip]
        {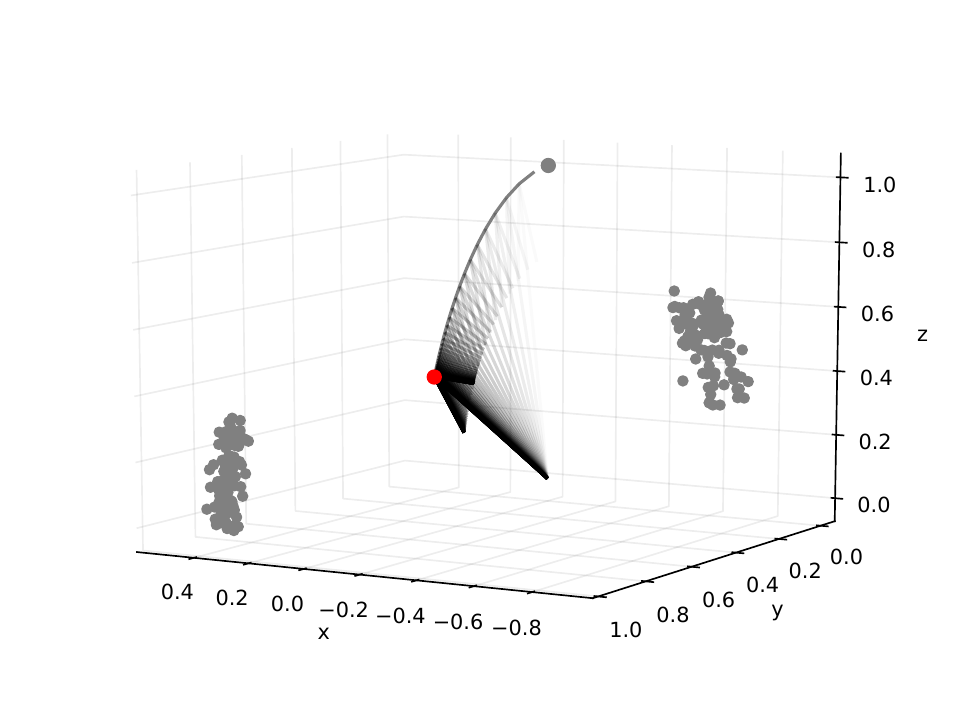}
        \caption{Without FoV limit.}
        \label{fig:mean}
    \end{subfigure}
    \hfill
    \begin{subfigure}[b]{0.48\columnwidth}
        \centering
        \includegraphics[width=\linewidth,trim={60pt 40pt 30pt 40pt},clip]
        {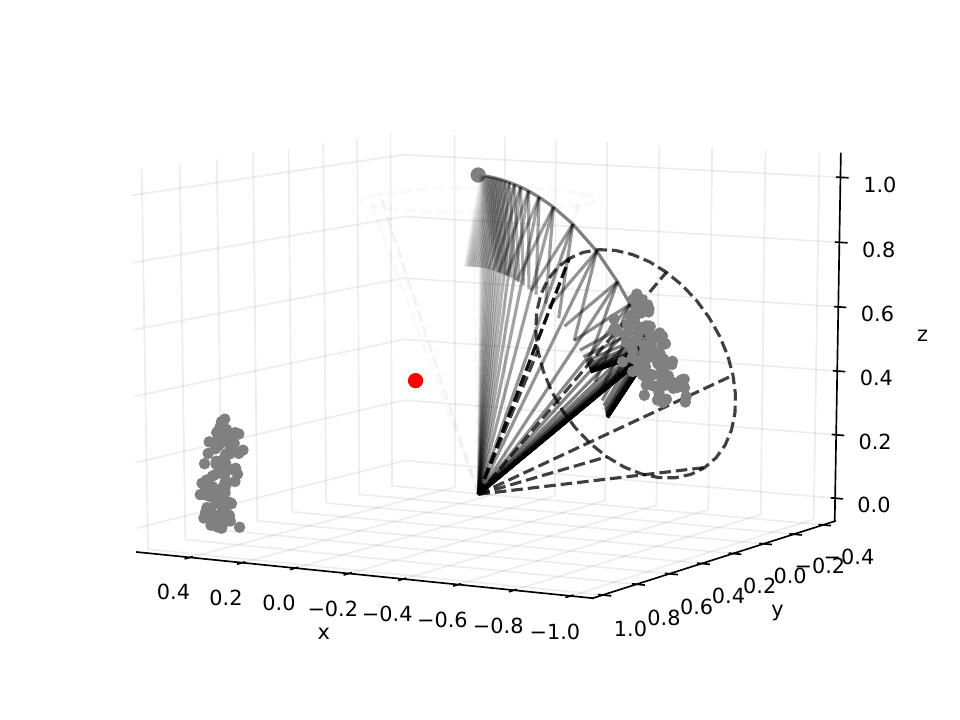}
        \caption{With $\alpha=45^\circ$.}
        \label{fig:FOV}
    \end{subfigure}
    \caption{Effect of the FoV constraint: (a) without an FoV limit, the
    optimized view points between the feature clusters; (b) with
    $\alpha=45^\circ$, it turns toward one cluster to maximize visible
    features.}
    \label{fig:fov_constraint_effect}
    \vspace{-10pt}
\end{figure}

\begin{figure*}[t]
\centering
\begin{subfigure}[t]{0.31\linewidth}
    \centering
    \includegraphics[
        trim=2.3cm 0.6cm 2cm 1cm,
        clip,
        width=\linewidth
    ]{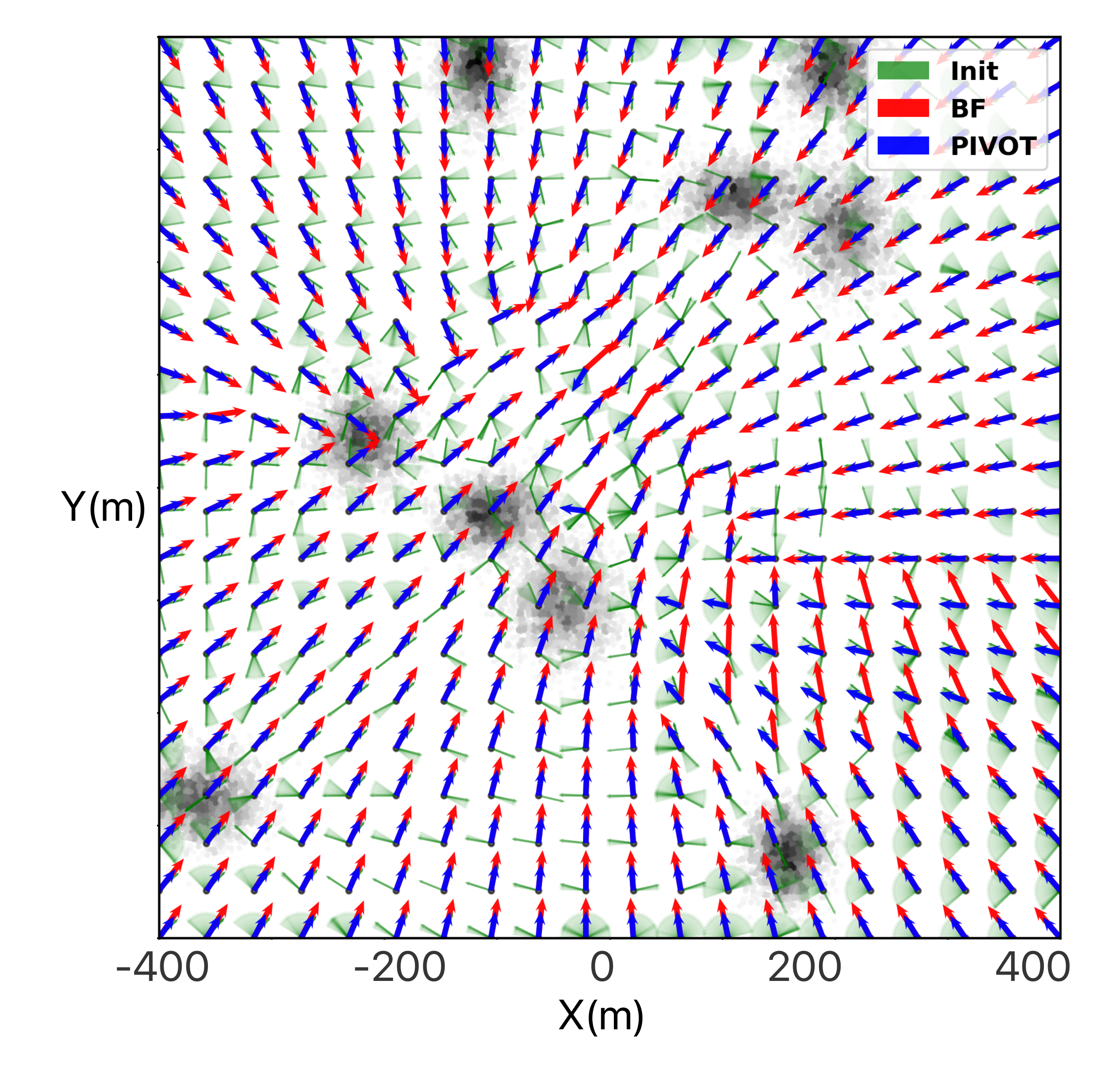}
    \caption{Viewing-direction comparison.}
    \label{fig:monte-orientations}
\end{subfigure}
\hfill
\begin{subfigure}[t]{0.26\linewidth}
    \centering
    \includegraphics[
        trim=7.2cm 1.1cm 4.2cm 1cm,
        clip,
        width=\linewidth
    ]{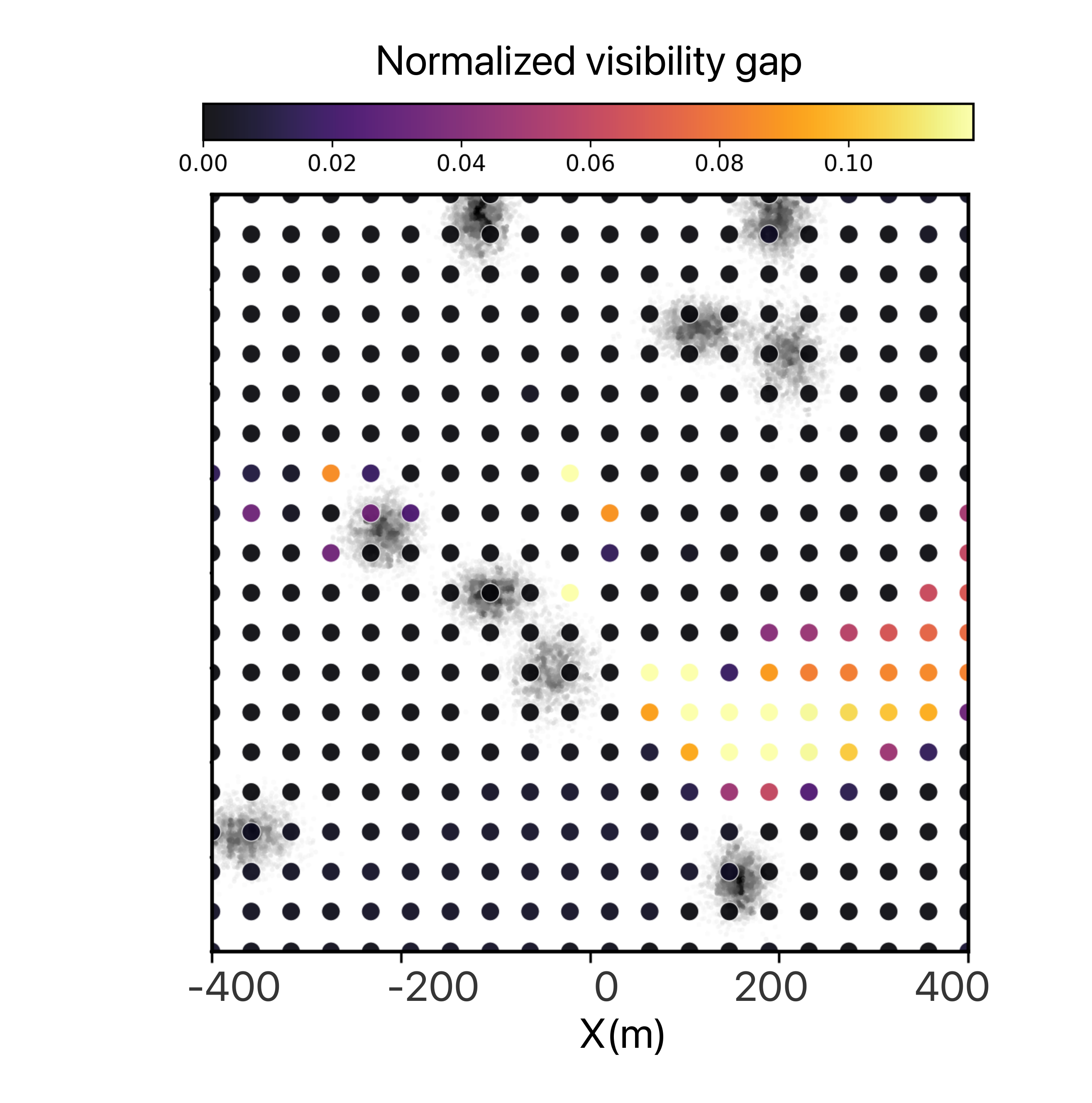}
    \caption{Normalized visibility gap.}
    \label{fig:monte-gap}
\end{subfigure}
\hfill
\begin{subfigure}[t]{0.40\linewidth}
    \centering
    \includegraphics[
        trim=1.8cm 1.2cm 2.7cm 1.5cm,
        clip,
        width=\linewidth
    ]{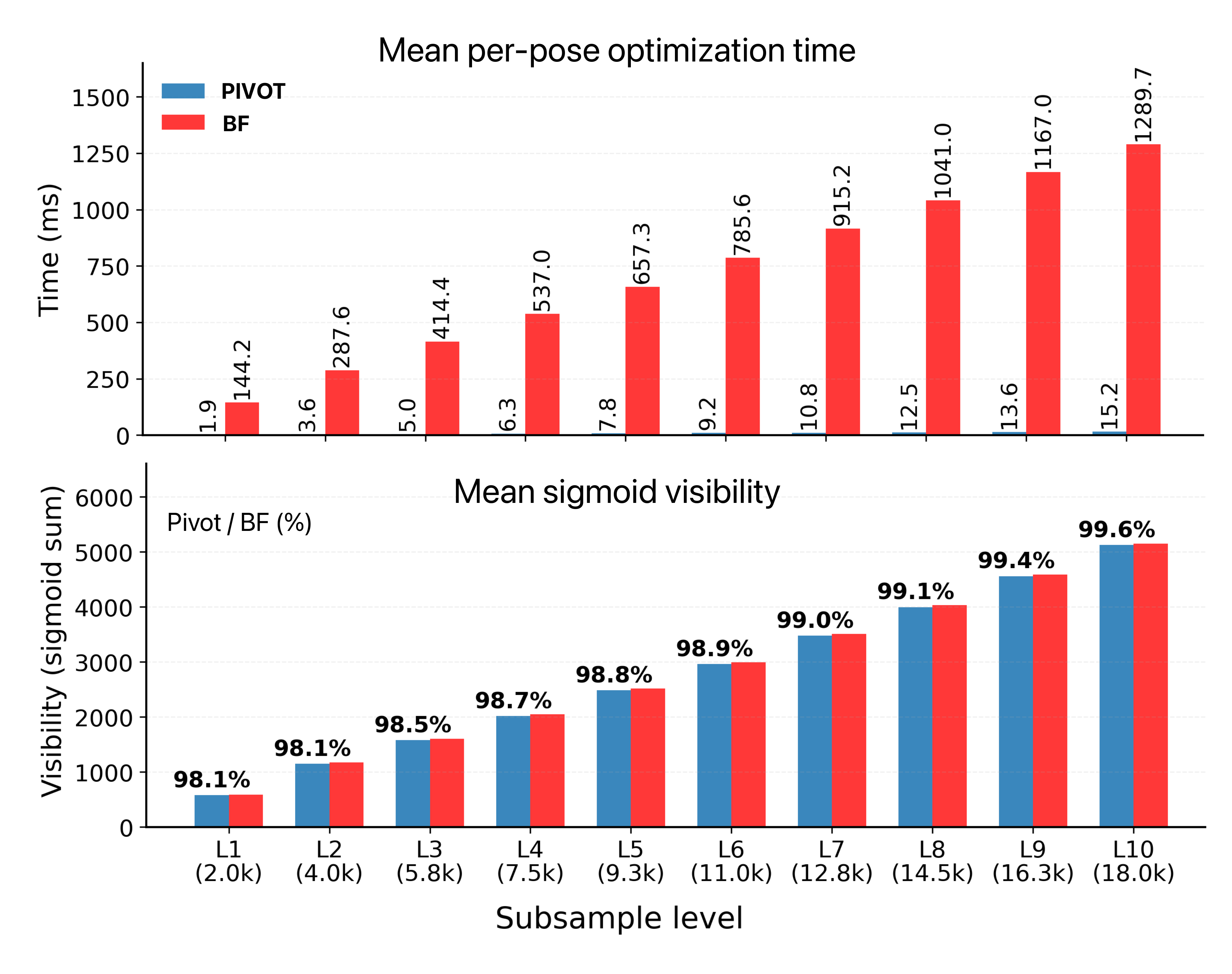}
    \caption{Scaling with feature count.}
    \label{fig:monte-runtime}
\end{subfigure}
\caption{Monte Carlo evaluation of the visibility--computation tradeoff.
Across spatially varying viewpoints and increasing map density, PIVOT
remains close to exhaustive search while scaling much more favorably
with feature count.}
\label{fig:monte-carlo-results}
\vspace{-12pt}
\end{figure*}

We evaluate the proposed method at three levels. First, we assess
single-pose viewing-direction optimization against a brute-force reference
(\mbox{\autoref{subsec:eval_visibility}}). Second, we examine
optimization along a fixed translation path to isolate the effect of
the trajectory smoothness objective
(\mbox{\autoref{subsec:eval_trajectory}}). Third, we evaluate
downstream visual localization in a photorealistic simulator
(\mbox{\autoref{subsec:eval_localization}}).
All timing experiments were performed on an Intel Core i9-14900K CPU
using single-threaded C++14 implementations under Ubuntu 22.04, with
wall-clock runtime measured using \texttt{std::chrono::steady\_clock}.

\subsection{Visibility Optimization}
\label{subsec:eval_visibility}

\subsubsection{Toy Example}

We first illustrate the effect of the FoV-aware objective using a
synthetic scene containing two clusters of 100 feature points.
Without an FoV constraint, maximizing summed cosine similarity to the
feature bearings aligns the optical axis with their aggregate direction
(\mbox{\autoref{fig:mean}}). With $\alpha=45^\circ$,
optimizing~\autoref{eq:objective} instead favors the cluster that
provides the larger aggregate visibility within the viewing cone
(\mbox{\autoref{fig:FOV}}). The smooth visibility approximation
in~\autoref{eq:smooth_visibility} provides a differentiable transition
at the FoV boundary, allowing continuous viewing-direction optimization
without discretizing the viewing sphere.

\subsubsection{Monte Carlo Evaluation}

We evaluate the proposed optimizer on synthetic maps generated from
nine truncated Gaussian clusters of 2000 features each, sampled within
$[-400,400]\times[-400,400]\times[0,20]$. Camera poses are placed on a
$20\times20$ grid spanning $x,y\in[-400,400]$ at fixed height $z=20$,
yielding 400 evaluation poses. Ten deterministic nested feature sets,
ranging from approximately 2k to 18k features, are used to evaluate
scaling with map density. The full FoV opening angle is $30^\circ$, and
all features use uniform weights $s_j=1$. As a reference, brute force
evaluates~\autoref{eq:objective} over the viewing sphere at $2^\circ$
angular resolution.

\par\noindent\textbf{Metrics.}
Viewpoint quality is measured relative to the brute-force reference
using the normalized visibility gap,
$(V_{\mathrm{BF}}-V_{\mathrm{PIVOT}})/V_{\mathrm{BF}}$, and retained
visibility, $100\,V_{\mathrm{PIVOT}}/V_{\mathrm{BF}}$.
Computational cost is measured by mean per-pose optimization time and
speedup relative to brute force.

\par\noindent\textbf{Results.}
\autoref{fig:monte-carlo-results} summarizes the results. The optimized
viewing directions closely match the brute-force solutions over most of
the pose grid (\mbox{\autoref{fig:monte-orientations}}), with deviations
occurring primarily where the optimizer converges to nearby local optima
associated with other feature clusters. The normalized visibility gap is
near zero for most poses, with larger discrepancies concentrated in a
small number of regions (\mbox{\autoref{fig:monte-gap}}). Across the ten
feature-count levels, our method requires 1.9--15.2\,ms per pose,
compared with 144.2--1289.7\,ms for brute force, corresponding to a
$76\times$--$85\times$ speedup while retaining 98.1\%--99.6\% of the
brute-force visibility (\mbox{\autoref{fig:monte-runtime}}).

\subsection{Trajectory Optimization}
\label{subsec:eval_trajectory}

\begin{figure}[t]
    \centering
    \includegraphics[width=\linewidth,trim={37pt 35pt 30pt 35pt},clip]
    {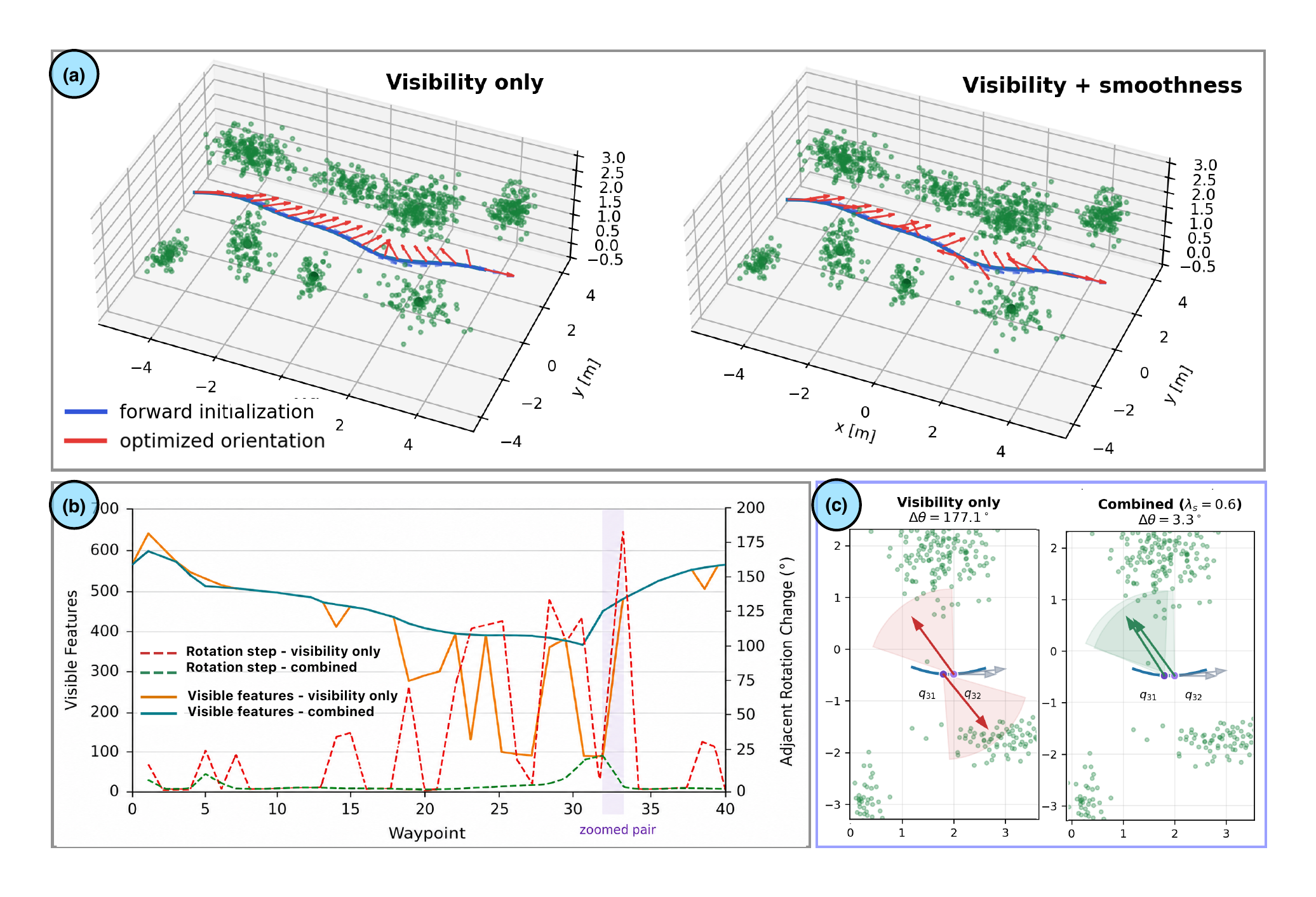}
    \caption{Effect of trajectory-level smoothness. (a) Visibility-only
    optimization produces abrupt viewpoint changes, while the combined
    objective yields smoother viewing directions along the same fixed
    path. (b) Visible-feature count and adjacent-view angular change along
    the trajectory. The highlighted pair is enlarged in (c), where the
    viewing-direction change decreases from $177.1^\circ$ to $3.3^\circ$.}
    \label{fig:traj-smoothness-comparison}
    \vspace{-15pt}
\end{figure}

We evaluate the trajectory-level smoothness objective on a fixed
RRT$^\ast$ translation path through multiple feature clusters. Sensor
orientations are initialized such that their optical axes align with the
local path direction and are optimized either for visibility alone
($\lambda=0$) or with the combined visibility and smoothness objective
($\lambda=0.6$).

\par\noindent\textbf{Metrics.}
We report the hard visible-feature count at each waypoint and the
angular change between consecutive viewing directions,
$\Delta\theta_i=\cos^{-1}((R_i^w\mathbf e_3)^\top
(R_{i+1}^w\mathbf e_3))$. These measure retained visibility and
viewing-direction smoothness, respectively.

\par\noindent\textbf{Results.}
As shown in~\autoref{fig:traj-smoothness-comparison}, visibility-only
optimization can switch between competing feature clusters, causing
large viewing-direction discontinuities despite high feature visibility.
The smoothness term largely preserves the visibility profile while
suppressing these abrupt changes. At the highlighted waypoint pair,
$\Delta\theta$ decreases from $177.1^\circ$ to $3.3^\circ$.

\subsection{Visual Localization}
\label{subsec:eval_localization}

We evaluate the proposed method in a visual localization setting using
the same NVIDIA Isaac / Unreal Engine photorealistic simulator
as~\cite{zhang2020fisher}. Following its mapping pipeline, images and
depth maps are rendered from the simulator and used, together with
camera poses, to build a sparse SfM map of SIFT features via
COLMAP~\cite{Schonberger_2016_CVPR} and an occlusion depth map from a
regular 3D voxel grid. We use two landmark maps: r1-a30 (1445 SIFT
features, $\alpha=30^\circ$) and r2-a20 (3470 SIFT features,
$\alpha=20^\circ$). The same SfM and depth maps are shared by all
methods.

\par\noindent\textbf{Baselines.}
We compare against the six perception-aware baselines
from~\cite{zhang2020fisher}: PC-D, PC-T, GP-D, GP-T,
Quad-D, and Quad-T, together with a no-information baseline.
Here, D and T denote the determinant- and trace-based Fisher
information objectives, respectively. PC-D and PC-T compute these
metrics directly from the landmark point cloud, whereas GP-D, GP-T,
Quad-D, and Quad-T precompute a Fisher information field
offline~\cite{zhang2020fisher} using GP- or quadratic-based visibility
approximations for efficient online queries. For the six
perception-aware baselines, perception is integrated into joint
translation and yaw planning, with sensor direction coupled to the
robot heading.

\begin{figure}[t]
\centering
\includegraphics[width=0.48\textwidth,trim={90pt 20pt 40pt 20pt},clip]{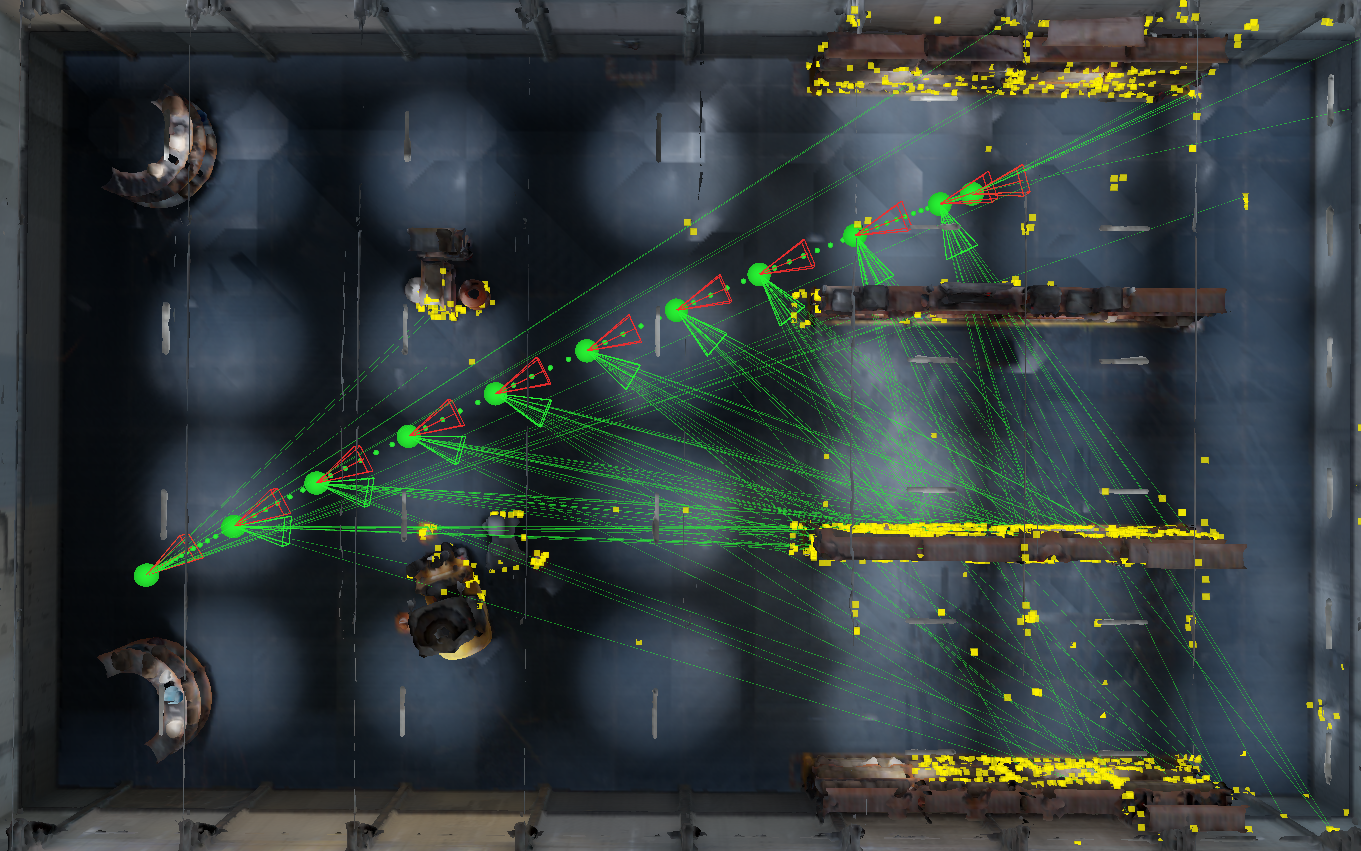}
\caption{Fixed translation path with original (red) and optimized (green) camera viewing directions. Green rays indicate visible SIFT features.}
\label{fig:example_plan}
\vspace{-5pt}
\end{figure}

\par\noindent\textbf{Our pipeline.}
Our method decouples translation planning from FoV optimization.
A collision-free translation path (xyz only) is first generated with
a piecewise polynomial planner similar to~\cite{zhang2020fisher},
using the same ESDF as the baselines for collision checking. Sensor
pointing is then optimized along the resulting path using the proposed
$SO(3)$ exponential-map updates. Following~\cite{zhang2020fisher}, we
use the prebuilt depth map to prefilter occluded landmarks at each
waypoint, yielding the retained landmark set $\mathcal{V}_i$. The
initial sensor orientation at each waypoint is chosen such that its
optical axis aligns with the local path direction. An example trajectory
with optimized viewing directions is shown in~\autoref{fig:example_plan}.

\begin{table}[t]
  \centering
  \footnotesize
  \scriptsize
  \setlength{\tabcolsep}{4.5pt}
  \begin{tabular}{lrrrrrrrr}
    \toprule
    Map & No Info. & Quad-D & Quad-T & PC-D & PC-T & GP-T & GP-D & \textbf{PIVOT} \\
    \midrule
    r1-a30 & 65.4 & 63.7 & 41.0 & 66.0 & 58.9 & 43.0 & 53.3 & \textbf{29.6} \\
    r2-a20 & 17.4 & 28.6 & 8.4 & 17.4 & 9.3 & 4.4 & 8.4 & \textbf{2.4} \\
    \bottomrule
  \end{tabular}
  \caption{COLMAP registration failure rate (\%).}
  \label{tab:recal}
  \vspace{-10pt}
\end{table}

\par
\begin{figure}[t]
  \vspace{-5pt}
  \centering
  \includegraphics[
    width=0.48\textwidth,
    trim={40pt 40pt 50pt 20pt},
    clip
  ]{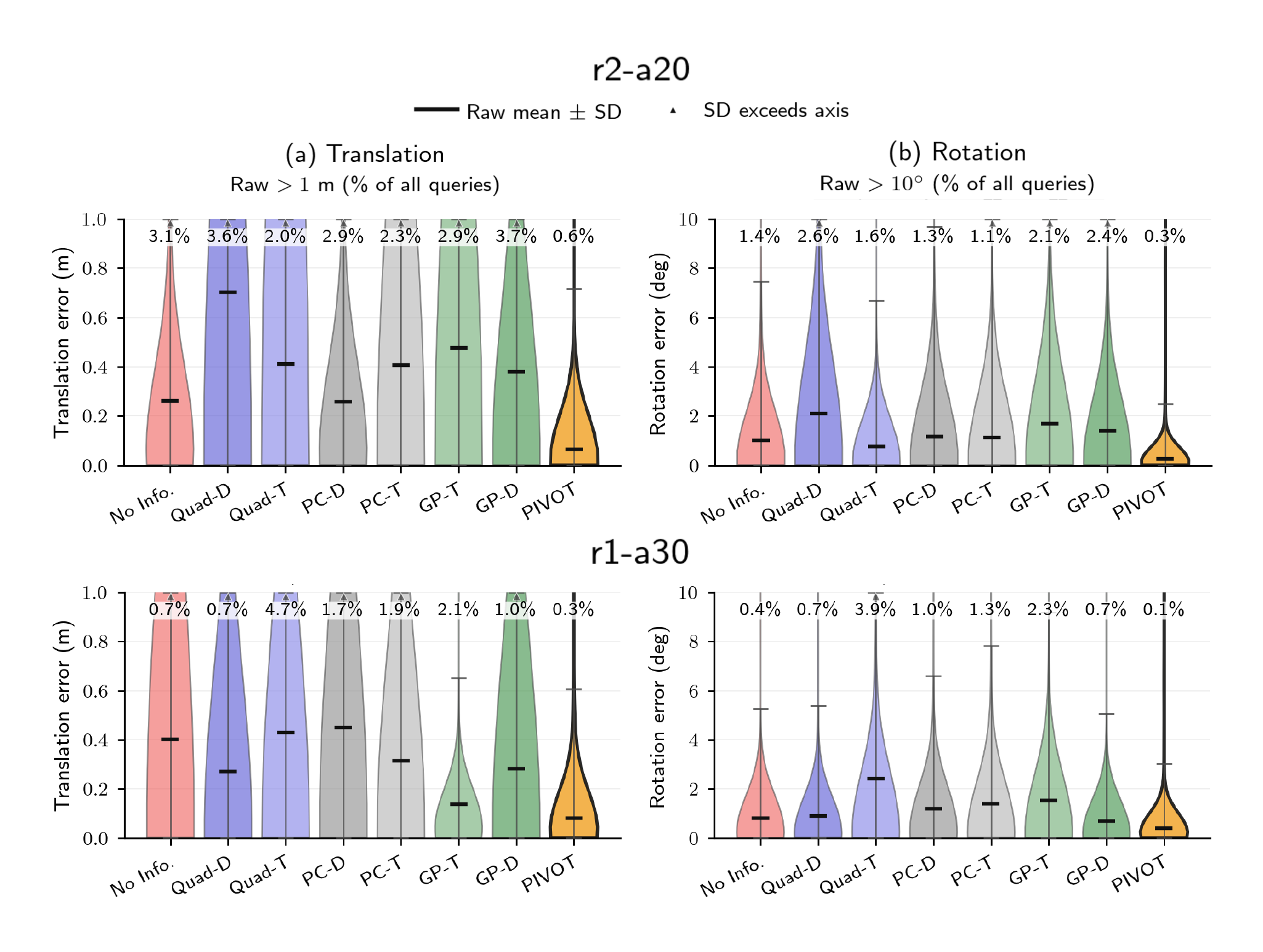}
  \caption{Pose-error distributions for r1-a30 (sparser map, wider FoV)
  and r2-a20 (denser map, narrower FoV).}
  \label{fig:violin}
\end{figure}

\noindent\textbf{Metrics.}
We evaluate computational efficiency using total optimization time and
localization performance using translation and rotation pose errors
together with COLMAP registration-failure rate. For each trajectory,
100 poses are rendered and registered against the SfM reference map in
COLMAP. The violin plots in~\autoref{fig:violin} use all finite pose
errors; the displayed axes are restricted to $0$--$1\,\mathrm{m}$ and
$0$--$10^\circ$ for readability. Percentages above the violins report
the fraction of all queries with finite errors beyond these limits.
Queries for which COLMAP returns no valid pose are excluded from the
violin distributions and reported separately as registration failures
in~\autoref{tab:recal}.

\par\noindent\textbf{Results.}
\textit{Computation time.}
\autoref{tab:runtime_two_maps_depthmap} reports the computation-time
results. For FIF-based baselines, offline field-construction time is
reported separately from online planning time. PIVOT requires
0.045\,s on r1-a30 and 0.061\,s on r2-a20 in total, compared with
5.6--41.8\,s for the baselines.

\textit{Localization accuracy.}
As shown in~\autoref{fig:violin}, PIVOT achieves the
lowest mean translation and rotation errors on both maps and the
smallest fractions of finite errors beyond the displayed limits.
The registration-failure rates in~\autoref{tab:recal} show the same
trend. PIVOT achieves the lowest failure rate on both maps: 29.6\% on
r1-a30 and 2.4\% on r2-a20, compared with 41.0\% and 4.4\% for the
respective next-best baselines. Overall, PIVOT achieves the best
localization accuracy and registration robustness among the evaluated
methods while requiring roughly two to three orders of magnitude less
total computation time.

\begin{table}[t]
\centering
\footnotesize
\setlength{\tabcolsep}{2.1pt}
\renewcommand{\arraystretch}{1.12}
\begin{tabular}{l c|c c c|c c c}
\hline
\multirow{2}{*}{Method}
& \multirow{2}{*}{FIF build}
& \multicolumn{3}{c|}{r1-a30}
& \multicolumn{3}{c}{r2-a20} \\
\cline{3-8}
&
& Traj. plan & FoV opt & Total
& Traj. plan & FoV opt & Total \\
\hline
GP-T
& 26.480 & 0.204 & 0 & 26.68
& 0.217 & 0 & 26.70 \\
GP-D
& 41.100 & 0.470 & 0 & 41.57
& 0.682 & 0 & 41.78 \\
Quad-T
& 5.553 & 0.085 & 0 & 5.64
& 0.084 & 0 & 5.64 \\
Quad-D
& 6.695 & 0.197 & 0 & 6.89
& 0.170 & 0 & 6.86 \\
PC-T
& 0 & 6.671 & 0 & 6.67
& 19.295 & 0 & 19.29 \\
PC-D
& 0 & 7.563 & 0 & 7.56
& 15.846 & 0 & 15.85 \\
\hline
\textbf{PIVOT}
& \textbf{0}
& \textbf{0.037} & \textbf{0.008} & \textbf{0.045}
& \textbf{0.044} & \textbf{0.017} & \textbf{0.061} \\
\hline
\end{tabular}
\caption{Mean computation time over the evaluated trajectories.
``FIF build'' denotes offline Fisher Information Field construction.
``Traj. plan'' denotes joint position-and-yaw planning for the
perception-aware baselines and translation-only planning for PIVOT.
``FoV opt'' denotes the subsequent sensor-pointing optimization used
only by PIVOT.}
\vspace{-12pt}
\label{tab:runtime_two_maps_depthmap}
\end{table}
\section{Evaluation in Real-World Experiments}
\label{sec:evalreal}
We evaluate PIVOT on a physical robot to assess real-world localization
and viewpoint adaptation.
\subsection{Experimental Platform and Setup}
\label{subsec:real_setup}
Our platform consists of a Boston Dynamics Spot
quadruped,\footnote{\url{https://bostondynamics.com/products/spot/}}
a ROS-based pan--tilt gimbal (iQuotient
Robotics),\footnote{\url{https://www.robotshop.com/products/iquotient-robotics-ros-based-pan-tilt}}
an Intel RealSense D455 camera,\footnote{\url{https://www.intelrealsense.com/depth-camera-d455/}}
and a Velodyne VLP-16 LiDAR providing metric localization through
FAST-LIO.

\begin{figure*}[t]
\centering
\includegraphics[width=0.99\linewidth,trim={1cm 2cm 1cm 0cm},clip]
{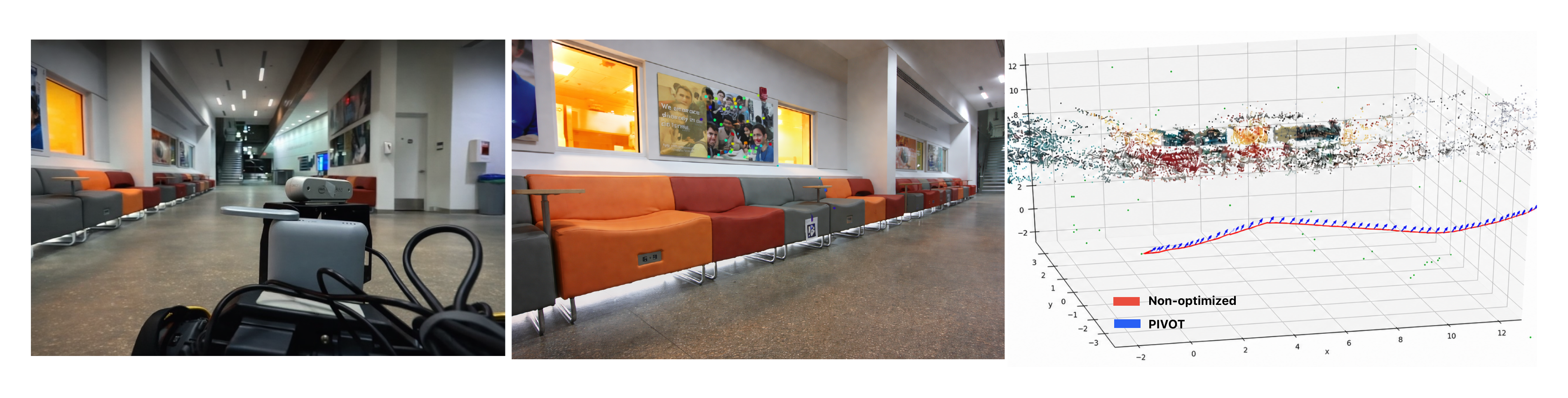}
\caption{Indoor localization experiment. Spot traverses the same
nominal Autowalk route with both forward-facing camera (red) and PIVOT
controlling the gimbal (blue). The middle panel shows SIFT detections in
the hallway, and the right panel illustrates the corresponding sensor
viewing directions along the route.}
\label{fig:indoors}
\end{figure*}
\begin{table}[t]
\centering
\setlength{\tabcolsep}{6pt}
\renewcommand{\arraystretch}{1.3}
\begin{tabular}{l c c}
\hline
Condition & Failures / Total & Success rate \\
\hline
PIVOT           & \phantom{0}4 / 420 & 99.0\% \\
Forward-facing  & 124 / 420          & 70.5\% \\
\hline
\end{tabular}
\caption{COLMAP registration results for the indoor Autowalk experiment.}
\vspace{-10pt}
\label{tab:real_results}
\end{table}
\par\noindent\textbf{Indoor localization experiment.}
The mapping sequence is chosen such that the dominant SIFT landmarks
are concentrated on one side of the hallway, creating a
visibility-sensitive localization setting in which sensor pointing
directly affects the availability of mapped features. The hallway is
first mapped using FAST-LIO. RGB images collected during a mapping
traversal are associated with FAST-LIO poses and used to construct a
sparse SIFT landmark map in COLMAP, which serves as the localization
reference. During the PIVOT traversal, live FAST-LIO localization is
used to select and track the corresponding planned sensor pointing
along the route. Spot traverses the same nominal pre-programmed
S-shaped Autowalk route once under each of two conditions: with the
camera fixed forward and with the gimbal tracking the PIVOT
viewing-direction trajectory. The setup and executed viewing directions are shown in~\autoref{fig:indoors}. Images from both traversals are
independently registered against the same COLMAP reference map. 
Registration success rate, defined as the fraction of query images for
which COLMAP returns a valid camera pose, is used as the primary indoor
metric.

\begin{figure}[t]
\centering
\includegraphics[width=0.48\textwidth,trim={0pt 0pt 0pt 0pt},clip]
{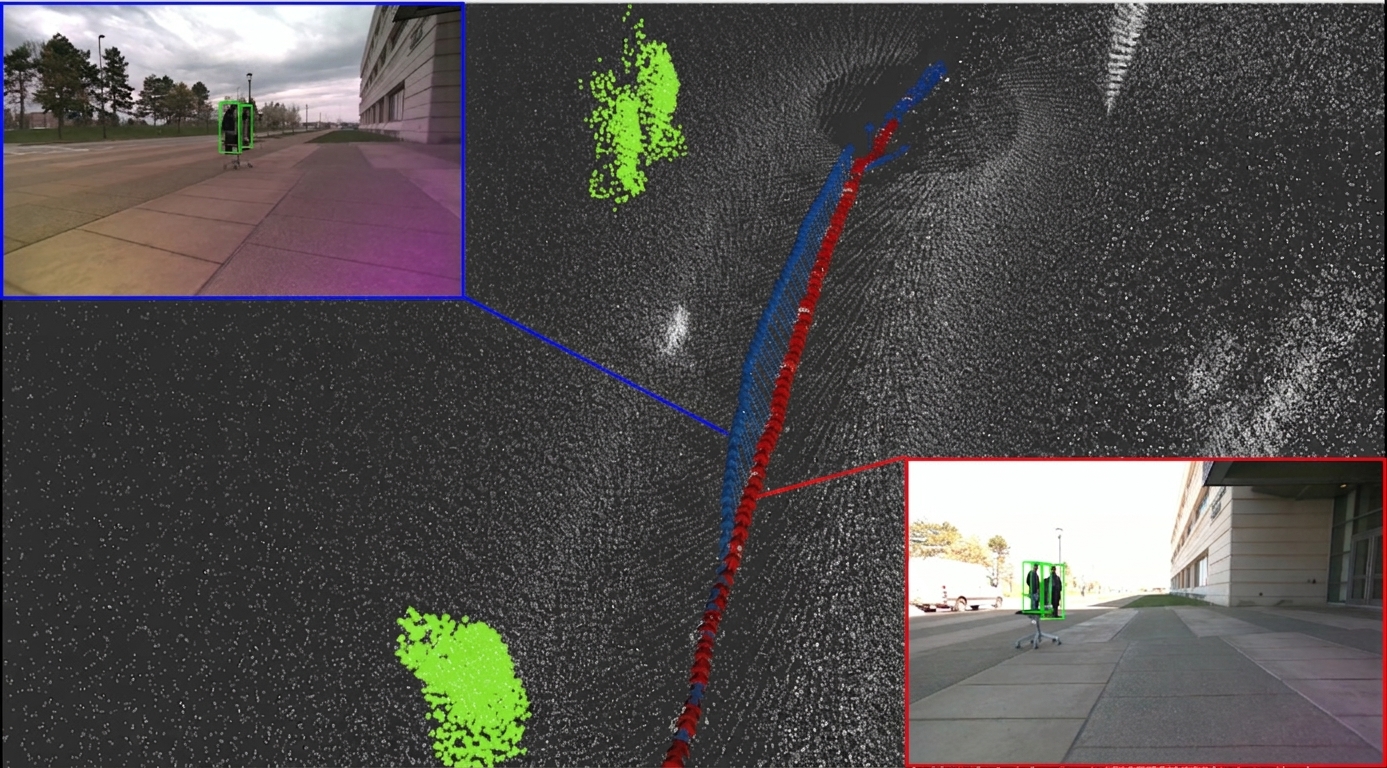}
\caption{Outdoor viewpoint-control demonstration; PIVOT targeting standing people as Spot traverses the scene.}
\vspace{-10pt}
\label{fig:outdoor}
\end{figure}

\par\noindent\textbf{Outdoor viewpoint-control experiment.}
We additionally evaluate semantically targeted viewpoint control using
standing people as regions of interest. Target regions are identified
from a recorded point cloud and provided to PIVOT as task-relevant
features for viewpoint planning. During the robot traversal, the
resulting viewing-direction trajectory is tracked using live
localization.

\subsection{Results}
\label{subsec:real_results}

\par\noindent\textbf{Indoor hallway.}
Across the two recorded traversals, PIVOT achieves a 99.0\%
registration success rate (416/420 frames), compared with 70.5\%
(296/420) for the forward-facing condition
(\autoref{tab:real_results}), corresponding to a 28.6 percentage-point
increase in registration success. The higher registration rate is
consistent with PIVOT maintaining mapped visual structure within the
limited camera FoV.

\par\noindent\textbf{Outdoor demonstration.}
\autoref{fig:outdoor} qualitatively demonstrates physical execution of
a semantically targeted viewing-direction trajectory while the robot
moves through the environment.


\section{Discussion}
\label{sec:discussion}

\par\noindent\textbf{Modularity and the decoupling tradeoff.}
PIVOT separates translation planning from sensor-pointing optimization,
enabling lightweight integration with existing navigation stacks.
However, viewpoint optimization cannot recover from a prescribed path
with no useful viewing direction. PIVOT is therefore best suited to
fixed or constrained trajectories with sufficient sensor-pointing
freedom.

\par\noindent\textbf{Online deployment and semantic integration.}
The real-world experiments evaluate physical execution of PIVOT
viewpoints using live localization, while the outdoor experiment uses
preprocessed semantic target regions. PIVOT is independent of how
task-relevant 3D features are generated and can therefore be paired
with an online detector, segmentation module, or incremental mapping
front end. The millisecond-scale optimization demonstrated in
\autoref{sec:evalsim} supports online integration in which the feature
set and corresponding viewing directions are updated as the robot moves.

\par\noindent\textbf{Limitations and future directions.}
PIVOT optimizes weighted feature visibility rather than geometric
information measures such as Fisher information. The distance weights
$s_{ij}$ could be augmented with task relevance, landmark uncertainty,
or information-theoretic quantities when richer perception objectives
are required. Because PIVOT operates directly on the current landmark
set without a separately precomputed perception-quality field, it could
also be combined with incremental mapping or receding-horizon planning
to adapt viewpoints as the map or task changes. The quantitative
real-world localization evaluation is limited to a single indoor route
with one traversal per sensing condition; broader evaluation across
environments, trajectories, and repeated trials remains future work.

\section{Conclusion}
\label{sec:conclusion}

We presented PIVOT, an on-manifold method for active FoV control of
motion-decoupled sensors along a fixed translation trajectory. PIVOT
optimizes sensor viewing direction using a closed-form visibility
gradient with $SO(3)$ exponential-map updates and trajectory-level
viewing-direction smoothness. It retains over 98\% of brute-force
visibility quality at much lower computational cost and outperforms
the evaluated Fisher-information-based baselines in localization
accuracy and registration robustness.

\bibliographystyle{IEEEtran}
\bibliography{bibtex/bib/IEEEexample}

@article{bajcsy1988active,
  title={Active perception},
  author={Bajcsy, Ruzena},
  journal={Proceedings of the IEEE},
  volume={76},
  number={8},
  pages={966--1005},
  year={1988},
  publisher={IEEE}
}

@article{indelman2015planning,
  title={Planning in the continuous domain: A generalized belief space approach for autonomous navigation in unknown environments},
  author={Indelman, Vadim and Carlone, Luca and Dellaert, Frank},
  journal={The International Journal of Robotics Research},
  volume={34},
  number={7},
  pages={849--882},
  year={2015},
  publisher={SAGE Publications Sage UK: London, England}
}

@inproceedings{zhang2019beyond,
  title={Beyond point clouds: Fisher information field for active visual localization},
  author={Zhang, Zichao and Scaramuzza, Davide},
  booktitle={2019 International Conference on Robotics and Automation (ICRA)},
  pages={5986--5992},
  year={2019},
  organization={IEEE}
}

@article{zhang2020fisher,
  author={Zhang, Zichao and Scaramuzza, Davide},
  title={Fisher information field: an efficient and differentiable map for perception-aware planning},
  journal={arXiv preprint arXiv:2008.03324},
  year={2020}
}

@inproceedings{isler2016information,
  title={An information gain formulation for active volumetric 3D reconstruction},
  author={Isler, Stefan and Sabzevari, Reza and Delmerico, Jeffrey and Scaramuzza, Davide},
  booktitle={2016 IEEE International Conference on Robotics and Automation (ICRA)},
  pages={3477--3484},
  year={2016},
  organization={IEEE}
}

@article{liu2020real,
  title={Real-time visual tracking of moving targets using a low-cost unmanned aerial vehicle with a 3-axis stabilized gimbal system},
  author={Liu, Xuancen and Yang, Yueneng and Ma, Chenxiang and Li, Jie and Zhang, Shifeng},
  journal={Applied Sciences},
  volume={10},
  number={15},
  pages={5064},
  year={2020},
  publisher={MDPI}
}

@incollection{jakobsen2005control,
  title={Control architecture for a {UAV}-mounted pan/tilt/roll camera gimbal},
  author={Jakobsen, Ole and Johnson, Eric},
  booktitle={Infotech@ Aerospace},
  pages={7145},
  year={2005}
}

@inproceedings{zou2006pan,
  title={A pan-tilt camera control system of {UAV} visual tracking based on biomimetic eye},
  author={Zou, Hairong and Gong, Zhenbang and Xie, Shaorong and Ding, Wei},
  booktitle={2006 IEEE International Conference on Robotics and Biomimetics},
  pages={1477--1482},
  year={2006},
  organization={IEEE}
}

@inproceedings{wang2019large,
  title={A large aperture 2-axis electrothermal {MEMS} mirror for compact 3D {LiDAR}},
  author={Wang, Dingkang and Watkins, Connor and Aradhya, Medini and Koppal, Sanjeev and Xie, Huikai},
  booktitle={2019 International Conference on Optical {MEMS} and Nanophotonics (OMN)},
  pages={180--181},
  year={2019},
  organization={IEEE}
}

@article{wang2020low,
  title={A low-voltage, low-current, digital-driven {MEMS} mirror for low-power {LiDAR}},
  author={Wang, Dingkang and Thomas, Lenworth and Koppal, Sanjeev and Ding, Yingtao and Xie, Huikai},
  journal={IEEE Sensors Letters},
  volume={4},
  number={8},
  pages={1--4},
  year={2020},
  publisher={IEEE}
}

@inproceedings{wang2017ultra,
  title={An ultra-fast electrothermal micromirror with bimorph actuators made of copper/tungsten},
  author={Wang, Dingkang and Zhang, Xiaoyang and Zhou, Liang and Liang, Mengyue and Zhang, Daihua and Xie, Huikai},
  booktitle={2017 International Conference on Optical {MEMS} and Nanophotonics (OMN)},
  pages={1--2},
  year={2017},
  organization={IEEE}
}

@article{costante2016perception,
  title={Perception-aware path planning},
  author={Costante, Gabriele and Forster, Christian and Delmerico, Jeffrey and Valigi, Paolo and Scaramuzza, Davide},
  journal={arXiv preprint arXiv:1605.04151},
  year={2016}
}

@inproceedings{alzugaray2017short,
  title={Short-term {UAV} path-planning with monocular-inertial {SLAM} in the loop},
  author={Alzugaray, Ignacio and Teixeira, Lucas and Chli, Margarita},
  booktitle={2017 IEEE international conference on robotics and automation (ICRA)},
  pages={2739--2746},
  year={2017},
  organization={IEEE}
}

@inproceedings{zhang2018perception,
  title={Perception-aware receding horizon navigation for {MAV}s},
  author={Zhang, Zichao and Scaramuzza, Davide},
  booktitle={2018 IEEE International Conference on Robotics and Automation (ICRA)},
  pages={2534--2541},
  year={2018},
  organization={IEEE}
}

@article{nageli2017real,
  title={Real-time motion planning for aerial videography with dynamic obstacle avoidance and viewpoint optimization},
  author={N{\"a}geli, Tobias and Alonso-Mora, Javier and Domahidi, Alexander and Rus, Daniela and Hilliges, Otmar},
  journal={IEEE Robotics and Automation Letters},
  volume={2},
  number={3},
  pages={1696--1703},
  year={2017},
  publisher={IEEE}
}

@article{gemerek2022directional,
  author  = {Jake Gemerek and Bo Fu and Yucheng Chen and Zeyu Liu and Min Zheng and David E. J. van Wijk and Silvia Ferrari},
  title   = {Directional Sensor Planning for Occlusion Avoidance},
  journal = {IEEE Transactions on Robotics},
  volume  = {38},
  number  = {6},
  pages   = {3713--3733},
  year    = {2022},
  month   = dec,
  doi     = {10.1109/TRO.2022.3180628}
}

@inproceedings{watterson2018trajectory,
  title={Trajectory Optimization On Manifolds with Applications to SO (3) and R3XS2.},
  author={Watterson, Michael and Liu, Sikang and Sun, Ke and Smith, Trey and Kumar, Vijay},
  booktitle={Robotics: Science and Systems},
  year={2018}
}

@InProceedings{Schonberger_2016_CVPR,
author = {Schonberger, Johannes L. and Frahm, Jan-Michael},
title = {Structure-From-Motion Revisited},
booktitle = {Proceedings of the IEEE Conference on Computer Vision and Pattern Recognition (CVPR)},
month = {June},
year = {2016}
}

@ARTICLE{chen2024design,
  author={Chen, Yuyang and Wang, Dingkang and Thomas, Lenworth and Dantu, Karthik and Koppal, Sanjeev J.},
  journal={IEEE Transactions on Robotics}, 
  title={Design of an Adaptive Lightweight {LiDAR} to Decouple Robot–Camera Geometry}, 
  year={2024},
  volume={40},
  number={},
  pages={2254-2271},
  doi={10.1109/TRO.2024.3371885}}

@article{tasneem2020adaptive,
  title={Adaptive fovea for scanning depth sensors},
  author={Tasneem, Zaid and Adhivarahan, Charuvahan and Wang, Dingkang and Xie, Huikai and Dantu, Karthik and Koppal, Sanjeev J.},
  journal={The International Journal of Robotics Research},
  volume={39},
  number={7},
  pages={837--855},
  year={2020},
  doi={10.1177/0278364920920931},
  url={https://doi.org/10.1177/0278364920920931}
}

@inproceedings{parandekar2024informative,
  title={Informative Sensor Planning for a Single-Axis Gimbaled Camera on a Fixed-Wing {UAV}},
  author={Parandekar, Aditya and Moon, Brady and Suvarna, Nayana and Scherer, Sebastian},
  booktitle={2024 IEEE 20th International Conference on Automation Science and Engineering (CASE)},
  pages={1798--1804},
  year={2024},
  organization={IEEE}
}

@article{wang2024active,
  author    = {Zhihao Wang and Haoyao Chen and Shiwu Zhang and Yunjiang Lou},
  title     = {Active View Planning for Visual {SLAM} in Outdoor Environments Based on Continuous Information Modeling},
  journal   = {IEEE/ASME Transactions on Mechatronics},
  volume    = {29},
  number    = {1},
  pages     = {237--248},
  year      = {2024},
  doi       = {10.1109/TMECH.2023.3272910}
}

@inproceedings{bonetto2021active,
  author    = {Elia Bonetto and Pascal Goldschmid and Michael J. Black and Aamir Ahmad},
  title     = {Active Visual {SLAM} with Independently Rotating Camera},
  booktitle = {2021 European Conference on Mobile Robots (ECMR)},
  pages     = {1--8},
  year      = {2021},
  organization = {IEEE},
  doi       = {10.1109/ECMR50962.2021.9568791}
}

@article{zhou2021raptor,
  author    = {Boyu Zhou and Jie Pan and Fei Gao and Shaojie Shen},
  title     = {{RAPTOR}: Robust and Perception-Aware Trajectory Replanning for Quadrotor Fast Flight},
  journal   = {IEEE Transactions on Robotics},
  volume    = {37},
  number    = {6},
  pages     = {1992--2009},
  year      = {2021}
}

@INPROCEEDINGS{patel2019,
  author={Patel, Bhavit and Warren, Michael and Schoellig, Angela},
  booktitle={2019 16th Conference on Computer and Robot Vision (CRV)}, 
  title={Point Me In The Right Direction: Improving Visual Localization on {UAV}s with Active Gimballed Camera Pointing}, 
  year={2019},
  volume={},
  number={},
  pages={105-112},
  doi={10.1109/CRV.2019.00022}}

@article{papaioannou2023integrated,
  author  = {Papaioannou, Savvas and Kolios, Panayiotis and
             Theocharides, Theocharis and Panayiotou, Christos G. and
             Polycarpou, Marios M.},
  title   = {Integrated Guidance and Gimbal Control for Coverage Planning
             With Visibility Constraints},
  journal = {IEEE Transactions on Aerospace and Electronic Systems},
  year    = {2023},
  volume  = {59},
  number  = {2},
  pages   = {1276--1291},
  doi     = {10.1109/TAES.2022.3199196}
}

@article{spurny2022active,
  author    = {Vojt{\v{e}}ch Spurn{\'y} and V{\'i}t Pritzl and Patrik B{\v{e}}lohradsk{\'y} and Martin Saska},
  title     = {Active Perception for {MAV}s With a Gimbal-Stabilized Camera},
  journal   = {IEEE Robotics and Automation Letters},
  volume    = {7},
  number    = {2},
  pages     = {3325--3332},
  year      = {2022},
  publisher = {IEEE}
}

@InProceedings{li2025actloc,
  title     = {{ActLoc}: Learning to Localize on the Move via Active Viewpoint Selection},
  author    = {Li, Jiajie and Sun, Boyang and Di Giammarino, Luca and Blum, Hermann and Pollefeys, Marc},
  booktitle = {Proceedings of The 9th Conference on Robot Learning},
  pages     = {1225--1245},
  year      = {2025},
  volume    = {305},
  series    = {Proceedings of Machine Learning Research}
}

@article{zhao2022perception,
  author    = {Yao Zhao and Zhi Xiong and Shuailin Zhou and Jingqi Wang and Ling Zhang and Pascual Campoy},
  title     = {Perception-Aware Planning for Active {SLAM} in Dynamic Environments},
  journal   = {Remote Sensing},
  volume    = {14},
  number    = {11},
  pages     = {2584},
  year      = {2022},
  publisher = {MDPI},
  doi       = {10.3390/rs14112584}
}

\end{document}